\documentclass[conference]{IEEEtran}
\IEEEoverridecommandlockouts

\usepackage{cite}
\usepackage{amsmath,amssymb,amsfonts}
\usepackage{algorithmic}
\usepackage{graphicx}
\usepackage{textcomp}
\usepackage{xcolor}
\def\BibTeX{{\rm B\kern-.05em{\sc i\kern-.025em b}\kern-.08em
    T\kern-.1667em\lower.7ex\hbox{E}\kern-.125emX}}

\newcommand{\ourmethod}{\textsc{PaSta}\xspace}
\usepackage{algorithmic}
\usepackage[linesnumbered,ruled,vlined]{algorithm2e}

\usepackage{threeparttable}
\usepackage{multirow}
\usepackage{rotating}
\usepackage{makecell}

\usepackage{amsmath}
\usepackage{amsthm}

\usepackage{amsfonts}

\usepackage{flushend}
\usepackage{color}
\usepackage{bm}
\usepackage{subfigure}

\usepackage{bbm}
\usepackage{amsthm}
\usepackage{graphicx}
\usepackage{subcaption}
\usepackage{amsmath}
\usepackage{amsfonts}

\usepackage{multirow} 
\usepackage{booktabs}
\newcommand{\eg}{{\em e.g.}}     
\newcommand{\ie}{{\em i.e.}}      

\usepackage{enumitem}
\usepackage{pifont}
\usepackage{url}

\newcommand{\eat}[1]{}

\begin{document}

\title{\ourmethod: Noisy Node Classification with \\ Partial Label Learning
}

\author{
\IEEEauthorblockN{
Yujing Liu,
Yixin Liu,
Yu Zheng,
Yue Tan,
Alan Wee-Chung Liew,
Shirui Pan$^*$ \thanks{$^*$ Shirui Pan is the corresponding author.}
}
\IEEEauthorblockA{
School of Information and Communication Technology, Griffith University, Brisbane, Australia\\
\{yujing.liu, yixin.liu, yu.zheng, yue.tan, a.liew, s.pan\}@griffith.edu.au
}
}

\maketitle

\begin{abstract}
Noisy node classification problem is a fundamental yet challenging task for real-world graph-related web services, where node labels are often corrupted or unreliable due to weak supervision or automatic annotation. 
However, existing methods typically train models based on one-hot labels, which not only makes models susceptible to overfitting on noisy labels, but also leads to error accumulation after pseudo-label-guided enhancement.  
In this paper, we propose a novel \textbf{Pa}rtial label-based \textbf{S}elf-\textbf{t}raining fr\textbf{a}mework (\ourmethod for short) that leverages partial label learning technique to overcome the limitations of existing methods. Specifically, \ourmethod first trains multiple annotators to comprehensively capture the class distribution of nodes and aggregates their predictions to construct high-quality partial labels. Subsequently, we design a partial label-based classification model with two well-crafted loss functions to guide the model learning at both label and representation spaces. 
To further enhance the robustness against noisy labels, we introduce a self-training strategy where the labels refined by partial label learning are then used to further optimize the annotators in a closed-loop iterative manner. 
Extensive experiments on five datasets demonstrate that, compared with existing state-of-the-art methods, \ourmethod achieves an average improvement of 1.1\% in classification performance under various noise settings. \textit{The source code is at \url{https://github.com/Yujingcn/PaSta-code}.}
\end{abstract}


\section{Introduction}
With the widespread application of graph-structured data in real-world web services such as social networks \cite{myers2014information}, recommendation systems \cite{wu2022graph}, and financial networks \cite{wang2022review}, node classification has become a key technique for tasks like user profiling \cite{purificato2023leveraging}, automatic product categorization \cite{yi2024research}, and fraud detection \cite{liu2020alleviating}. However, real-world graph data inevitably contains erroneous labels (\ie, noisy labels) due to factors such as manual annotation errors or delays in information updates. Existing node classification methods are prone to overfitting these noisy labels in practice, resulting in a significant drop in generalization performance~\cite{zhang2020adversarial}. Consequently, the problem of \textbf{\textit{noisy node classification}} (NNC) has gradually emerged as a central challenge for achieving reliable analysis and efficient utilization of graph-structured data in practical applications, attracting increasing research attention~\cite{liu2025noisy,lu2024noise2}.

\begin{figure}[t!]
    \centering
    \includegraphics[width=0.4\textwidth,]{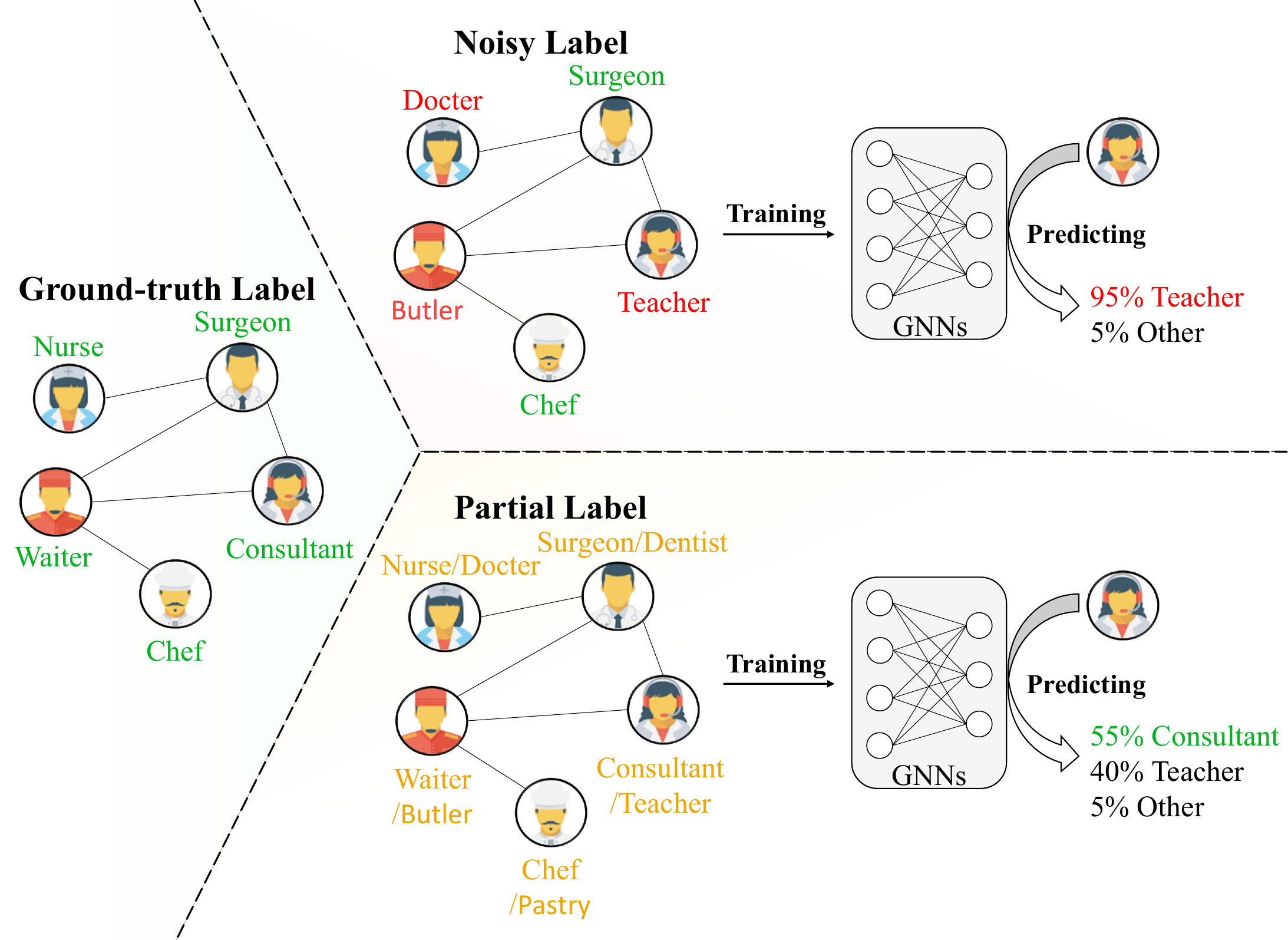}
    \caption{Sketch maps of noisy label and partial label learning. While noisy labels tend to drive the model toward confident but potentially incorrect predictions, partial labels help the model produce ambiguous yet more accurate predictions.}
    \label{fig:partial label}
\end{figure}

To address the NNC problem, researchers have proposed various algorithms to equip graph neural networks (GNNs) with enhanced techniques that improve robustness against noisy labels, such as graph structure optimization, consistency regularization, and label correction~\cite{liu2023multi,lu2024noise1,liu2025noisy,lu2024noise2}. Despite their effectiveness, the majority of current methods rely on one-hot labels to supervise the classifier, which can lead to two inherent limitations. \textit{\textbf{Limitation~1}:~Sensitivity to label noise}. Driven by the cross-entropy loss, one-hot labels always force the classifier to assign high-confidence predictions to a single category. Nevertheless, when the labels are noisy, the model is easily misled to fit the wrong labels, which significantly undermines both the performance and robustness of models~\cite{li2021unified,liu2023multi}. \textit{\textbf{Limitation~2}:~Difficulty in correcting label errors}. To address the NNC problem, existing methods often refine noisy labels or assign pseudo-labels to unlabeled nodes. 
However, pseudo-labels or refined labels may still be unreliable, and the absolute certainty imposed by one-hot label-guided training can further reinforce incorrect supervision and even amplify error accumulation~\cite{liu2025noisy}. Even though several methods adopt soft labels~\cite{li2021unified}, they are essentially probability-smoothed variants of one-hot labels (i.e., label smoothing), which still fail to capture the inherent uncertainty and ambiguity of labels.

Going beyond one-hot label modeling, \textbf{\textit{partial label learning}}~\cite{feng2019partial,saravanan2025pseudo} provides a promising paradigm to alleviate the above limitations. As shown in Figure~\ref{fig:partial label}, partial label learning relaxes the rigid one-hot assumption by allowing each sample to be annotated with multiple candidate labels, hence providing an effective way to model uncertainty and ambiguity in supervision. In this case, as long as the correct label is included in the candidate set, the model is able to learn the correct decision boundary~\cite{lv2024makes,sheng2024adaptive}, which substantially improves its robustness against noisy labels and thereby alleviates \textbf{\textit{Limitation 1}}. On the other hand, if we convert the pseudo-labels and refined labels into partial labels, the inherent uncertainty can be explicitly preserved, which prevents the model from being constrained by potentially incorrect one-hot supervision. In this way, \textbf{\textit{Limitation 2}} can be mitigated by reducing the risk introduced by faulty label correction.

Although partial label learning has shown great potential in handling noisy labels, how to apply this advanced paradigm to the NNC problem still remains non-trivial. Specifically, two critical questions remain to be solved: \textit{\textbf{Question~1}:~How to obtain reliable partial labels from graph data?} Unlike in computer vision domains, where high-quality partial labels can often be naturally available~\cite{wang2022pico,lv2020progressive} or easily acquired from pre-trained models (e.g., CLIP~\cite{radford2021learning}), it is difficult to construct reliable partial labels for real-world graph-structured data. Without high-quality partial labels, the classifier will be misled by vague and low-informative supervision signals, resulting in blurred decision boundaries and even model collapse. Hence, a robust annotation strategy is required to generate reliable partial label that are both complete and accurate. 
Even though accurate partial labels can be obtained, we have to face a follow-up challenge: \textit{\textbf{Question~2}:~How to effectively leverage partial labels to learn a GNN-based classifier?} Unlike one-hot labels that convey a probabilistic meaning with absolute certainty, partial labels characterize the possibility of multiple candidate classes and reflect the inherent ambiguity of supervision. In this case, directly applying the conventional cross-entropy loss for partial label learning may dilute the informative signal and ultimately hinder effective classifier training. Consequently, it is crucial to design learning objectives that can fully exploit the signals in partial labels.

In response to the above two questions, in this paper, we propose a \textbf{Pa}rtial label-based \textbf{S}elf-\textbf{t}raining fr\textbf{a}mework (\textbf{\ourmethod} for short) to tackle the NNC problem. To answer \textbf{\textit{Question~1}}, we introduce a \textit{self-supervised ensemble annotation module} to generate high-quality partial labels automatically. Specifically, we construct multiple annotators using different self-supervised learning strategies to pre-train their feature extractors, followed by learnable predictors. In this way, the diverse annotators can assign labels to nodes from different perspectives, and by ensembling them, we can acquire partial labels with accuracy and completeness. To address \textbf{\textit{Question~2}}, we develop a \textit{dual-space partial label learning module}, where a node classifier is trained with two complementary loss functions to effectively exploit partial labels. At label space, we design a vote-aggregated Cross Entropy (VaCE) loss, which leverages the partial labels obtained by aggregating votes from multiple annotators to optimize the output class probabilities, thereby explicitly leveraging the consensus distribution formed by annotators. At the latent representation space, we further propose a partial label similarity (PaSim) loss, which constrains node representations based on the pairwise similarity of their partial labels. The PaSim loss encourages nodes with similar candidate label sets to be closer in the representation space, enhancing the model’s ability to discriminate under label uncertainty. Furthermore, to improve robustness against noisy supervision in NNC, we employ a \textit{self-training strategy} that iteratively optimizes the annotation module and the partial label learning module in a bootstrapping manner. In each iteration, we enhance the labels for annotator training using the predictions from the partial label learning module; the refined partial labels, in turn, provide more reliable supervision to further improve the classifier. 
To sum up, the main contributions of our method are summarized as follows:

\begin{itemize}[leftmargin=*]
\item[$\bullet$] \textbf{New Paradigm.} To the best of our knowledge, we make the first attempt to use partial labels to address the NNC problem, which provides a new learning paradigm that explicitly models label uncertainty and enhances robustness against label noise. 

\item[$\bullet$] \textbf{Novel Method.} We propose \ourmethod, a self-training framework for NNC, which consists of a self-supervised ensemble annotation module and a dual-space partial label learning module. The two modules are jointly optimized in an iterative manner to mutually enhance partial label annotation quality and classifier performance.

\item[$\bullet$] \textbf{Ample Experiments.} We conduct extensive experiments on five real-world datasets, and the results show our method outperforms all existing SOTA methods across all noise levels.
\end{itemize}

\section{Related Works}
\subsection{Graph Neural Networks}
Graph Neural Networks (GNNs) have received widespread research attention due to their powerful capability in capturing structural topology and node attributes through recursive message-passing mechanisms. Consequently, GNNs are increasingly applied to a broad spectrum of downstream tasks, ranging from fundamental node classification \cite{shen2026raising, dai2021nrgnn} and graph anomaly detection \cite{liu2026few, liu2026rethinking, li2026towards1} to complex topology design \cite{li2026ofa}. Alongside this broadening scope of applications, advanced learning paradigms have developed rapidly to address real-world deployment challenges; for instance, federated graph learning frameworks \cite{tan2026influence, tan2023taming} manage data heterogeneity across distributed clients, while robust learning techniques \cite{yuan2023learning, liu2023multi} aim to mitigate noise interference during model training. Despite these broad successes, standard GNN architectures remain inherently vulnerable to corrupted graph data, as bad signals can easily propagate and compound across neighboring nodes. As a result, achieving robust model performance on real-world noisy graph data remains a pressingly unresolved problem that warrants further investigation.

\subsection{Noisy Node Classification}
GNNs are inherently sensitive to label noise because the recursive message-passing mechanism tends to propagate and amplify label errors across neighboring nodes, severely undermining model generalization \cite{kipf2017semi, dai2021say}. To address this, various robust GNN strategies have been proposed. Graph structure optimization methods, such as NRGNN \cite{dai2021nrgnn}, refine edge connections based on node similarity to ensure noisy nodes are supervised by reliable neighbors. Multi-teacher self-training frameworks (\eg MTS-GNN \cite{liu2023multi} and BO-NNC \cite{liu2025noisy}) utilize historical model states or distillation strategies to provide more stable supervision and guide label correction. Additionally, consistency-based models like CGCN \cite{yuan2023learning} leverage data augmentation and contrastive constraints to encourage the model to learn noise-invariant representations. Despite these advancements, existing methods often struggle with overfitting to noisy signals during late-stage training, and ensuring the absolute reliability of label correction remains an open challenge that \ourmethod seeks to overcome.

\subsection{Partial Label Learning}
Partial Label Learning (PLL) addresses scenarios where each instance is associated with a candidate label set containing the ground-truth \cite{gong2021top}. Traditional PLL algorithms focus on progressive disambiguation to identify the true label from uncertain sets \cite{wang2022pico, lv2020progressive}. Recently, PLL has been successfully adapted for noisy image classification (e.g., PALS \cite{saravanan2025pseudo} and NPN \cite{sheng2024adaptive}), where noisy labels are treated as candidates and refined through iterative optimization. However, the application of PLL to the NNC problem remains unexplored, primarily due to the difficulty of constructing high-quality candidate sets from noisy graph data, a gap that our proposed \ourmethod framework effectively fills.

\section{Preliminary}

\noindent\textbf{Notations.} 
A graph can be represented by $\mathcal{G}=(\mathcal{V}, \mathcal{E})$, where $\mathcal{V}=\{v_{1},\cdots,v_{n}\}$ denotes the node set with $n$ nodes, and $\mathcal{E}\subseteq \mathcal{V}\times \mathcal{V}$ corresponds to the edge set that encodes their pairwise relations. Each node $v_i$ is associated with a feature vector $\mathbf{x}_i \in \mathbb{R}^{d}$, and assembling all node features yields the feature matrix $\mathbf{X}=\{\mathbf{x}_1,\cdots,\mathbf{x}_n\}\in \mathbb{R}^{n\times d}$. The structure of the graph is represented by an adjacency matrix $\mathbf{A}\in\mathbb{R}^{n\times n}$, where each entry $\mathbf{a}_{ij}=1$ if $(v_i,v_j)\in \mathcal{E}$ and $\mathbf{A}_{ij}=0$ otherwise. 

\noindent\textbf{Noisy node classification (NNC).} 
In the NNC setting, only a small portion of nodes $\mathcal{V}_L \subseteq \mathcal{V}$ are provided with labels, denoted as $\mathbf{Y}=\{\mathbf{y}_1,\cdots,\mathbf{y}_b\}$, where $b=|\mathcal{V}_L|$ is the number of labeled nodes. Each label $\mathbf{y}_i \in \mathbb{R}^c$ is expressed as a one-hot vector over $c$ possible classes. Given $\mathcal{G}=(\mathcal{V},\mathcal{E})$ with feature matrix $\mathbf{X}$, adjacency matrix $\mathbf{A}$, and a noisy label set $\hat{\mathbf{Y}}$ (in which some annotations may be erroneous, i.e., $\hat{\mathbf{y}}_i \neq \mathbf{y}_i$ for certain $v_i \in \mathcal{V}_L$), the task of NNC is to develop a model that can accurately predict the ground-truth class of each node by exploiting $\hat{\mathbf{Y}}$ as supervisory signals.

\noindent\textbf{Partial label learning.} 
In partial label learning, each node $v_i \in \mathcal{V}_L$ is associated not with a single label but with a \emph{candidate label set} $Y_i \in \{0,1\}^C$, where the $j$-th element $Y_{ij}=1$ indicates that class $j$ is a candidate label and $Y_{ij}=0$ otherwise. The ground-truth label $y_i$ is guaranteed to be contained in $Y_i$, i.e., $y_i \in Y_i$, but its exact identity is unknown. The task of partial label learning is to develop a model that, given the candidate label set $Y_i$ for each labeled node $v_i \in \mathcal{V}_L$, can accurately infer the true class $y_i$, thereby learning classification patterns that generalize to all nodes $v \in \mathcal{V}$.



\section{Methodology}
\begin{figure*}[t!]
    \centering
    \includegraphics[width=0.8\textwidth,]{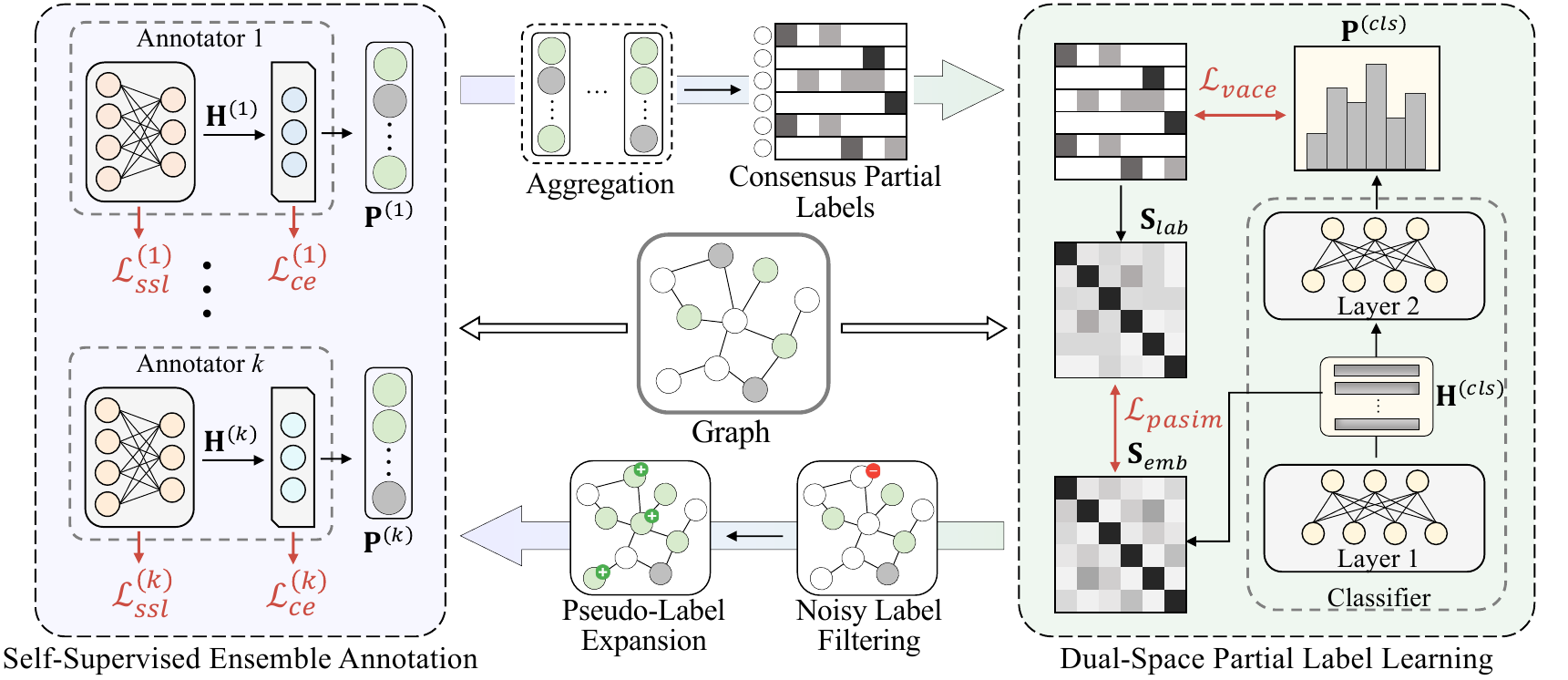}
    \caption{The framework of the proposed \ourmethod. On the left, we leverage diverse self-supervised graph learning algorithms to construct annotators and generate consensus labels. On the right, we introduce $\mathcal{L}_{pasim}$ and $\mathcal{L}_{vace}$ to guide the classifier in learning robust patterns from the consensus labels. The trained classifier is then used to filter and expand dataset labels, which are subsequently used to further optimize the annotators’ predictors. Finally, the two modules are iteratively optimized.}
    \label{fig:framework}
\end{figure*}



In this section, we introduce the proposed method, \textbf{Pa}rtial label-based \textbf{S}elf-\textbf{t}raining fr\textbf{a}mework (\textbf{\ourmethod}), for noisy node classification (NNC). With its overall pipeline illustrated in Figure~\ref{fig:framework}, \ourmethod is composed of two modules, namely \textit{self-supervised ensemble annotation} and \textit{dual-space partial label learning}. 
In the {self-supervised ensemble annotation module}, our method employs multiple self-supervised graph techniques to build diverse partial label annotators, enabling them to capture node category information more comprehensively and thereby generate partial labels with high correct information quantity. Subsequently, in the dual-space partial label learning module, we design two novel loss functions to guide the classifier in learning from partial labels, which enables \ourmethod to exploit the supervisory information at both label space and representation space. To further refine the annotated partial labels and classifier, we incorporate a self-training strategy that iteratively updates the two modules in a bootstrapping manner. To achieve this, we design a label enhancement process for noisy label filtering and pseudo-label expansion, which polishes the label information for annotator training. 



\subsection{Self-Supervised Ensemble Annotation}

To build a robust partial label learning model, constructing reliable partial labels is the cornerstone of the entire framework. While acquiring high-quality partial labels in other domains, such as computer vision and natural language processing, is relatively easy through multi-class annotations or large pre-trained models (e.g., CLIP or BERT \cite{devlin2019bert}), obtaining them in graph-structured data is far more challenging due to the absence of such annotation resources and the complexity of graph topology. In the context of NNC, the construction of partial labels becomes far more challenging since noisy labels can further contaminate the candidate label sets and degrade their overall quality. As a result, it is crucial to design a robust annotation strategy that can ensure the completeness and accuracy of partial labels.

To tackle this challenging task, we design a self-supervised ensemble annotation module, in which multiple annotators collaborate to generate partial labels together. For each annotator, we use a specialized self-supervised learning strategy to pre-train the feature extractor, which ensures diversity in the representation spaces among different annotators. Then, we train the predictors with enhanced labels (the enhancement process is detailed in Section~\ref{sec:enhance}), and the predictions from different annotators are finally ensembled to form high-quality consensus partial labels. These sub-components are introduced in the following paragraphs. 

\subsubsection{\textbf{Self-Supervised Feature Extractors}}
In order to construct diverse annotation sources for high-quality partial label construction, we first adopt multiple self-supervised graph learning methods to train the feature extractors of the annotators. Formally, for the $i$-th annotator, the learned node embeddings are learned from different perspectives:
\begin{eqnarray}
\mathbf{H}^{(i)} = \mathbf{F}_{enc}^{(i)}(\mathbf{X}, \mathbf{A}),
\quad \text{where } \mathbf{F}_{enc}^{(i)} = \arg\min_{\mathbf{F}} \mathcal{L}_{ssl}^{(i)}.
\label{eq3}
\end{eqnarray}
In this equation, $\mathbf{F}_{enc}^{(i)}(\cdot)$ denotes the feature extractor of $i$-th annotator, $i\in [1,..., t]$ with $t$ as the number of annotators, and $\mathcal{L}_{ssl}^{(i)}$ is a specialized self-supervised loss function to optimize $\mathbf{F}_{enc}^{(i)}(\cdot)$. The candidates of $\mathcal{L}_{ssl}^{(i)}$ can be mainstream graph self-supervised learning objectives, such as DGI~\cite{velivckovic2018deep}, GCA~\cite{zhu2021graph}, and SUGRL~\cite{mo2022simple}. 

The design of multiple self-supervised feature extractors has several advantages. \textit{Firstly}, since different graph self-supervised learning methods capture distinct structural or semantic properties, they can project nodes into diverse representation spaces, thereby ensuring the completeness of the constructed partial labels \cite{wu2022automated}. \textit{Secondly}, as self-supervised training does not rely on ground-truth labels, the learned extractors are immune to the influence of noisy labels, which satisfies the requirement of accuracy for partial labels. \textit{Thirdly}, self-supervised extractors can be pre-trained in advance during model initialization, which improves the training efficiency of \ourmethod.

\subsubsection{\textbf{Supervised Predictors}}
After the pre-training of feature extractors, for each annotator, we train a light-weight predictor (i.e., a fully connected layer with Softmax activation) via supervised learning. Specifically, the predictor is trained by the cross-entropy loss with a labeled node set:
\begin{eqnarray}
\mathcal{L}^{(i)}_{ce} = -\sum_{v_j \in \mathcal{S}}\sum_{k=1}^{c} \mathbb{I}(y_{jk} \neq 0) \log \mathbf{p}_{j}^{(i)},
\label{eq4}
\end{eqnarray}
where $\mathcal{S}$ denotes the labeled training set, and $\mathbf{p}_{j}^{(i)} \in \mathbb{R}^{c}$ and $\mathbf{y}_{j}$ denote the prediction of the predictor $\mathbf{F}_{pre}^{(i)}(\cdot)$ and the one-hot label of node $v_j$, respectively. Here, the training set $\mathcal{S}$ can be initialized as the original labeled set $\mathcal{V}_L$. As the model iteratively updates, $\mathcal{S}$ is progressively refined into enhanced labels (see Section~\ref{sec:enhance}). 

Although different predictors are trained with the same supervision signals, their distinct underlying representation spaces still lead them to exhibit diverse feature extraction and classification behaviors. Consequently, different annotators can still exhibit diverse classification behaviors, thereby differentiating the impact of noisy labels on them and making their predictions complementary. 

\subsubsection{\textbf{Ensembled into Consensus Partial Labels}}
Finally, the prediction results of these annotators are aggregated to form consensus partial labels. Concretely, the consensus label vector for node $v_i$ is given by collecting the most confident category from each annotator:
\begin{equation}
\tilde{\mathbf{y}}_i = \left[\tilde{y}_{i1}, \dots, \tilde{\mathbf{y}}_{ic}\right], \quad \text{where}\
\tilde{y}_{ij} = \sum_{k=1}^{t} \mathbb{I}\big(\arg\max \mathbf{p}_i^{(k)} = j\big).
\label{eq5}
\end{equation}

It is worth noting that, in \ourmethod, we construct a well-designed \textit{consensus partial labels}, rather than the traditional binarized partial labels, because we argue that non-binarized consensus labels can carry richer supervisory signals. Firstly, non-binarized labels can more precisely reflect the positional relationships of nodes in the partial label space, thereby providing more detailed inter-node semantic information for the classifier during subsequent training. Secondly, non-binarized consensus labels effectively serve as adaptive sample reweighting, guiding the model to prioritize nodes on which the annotators agree, which helps constrain the model to learn correct classification information from the remaining consensus labels rather than fitting error labels.

\subsection{Dual-Space Partial Label Learning}

Once we obtain the consensus partial labels, the following challenge is how to effectively leverage them to train a node classifier. While using a conventional training objective (e.g., cross-entropy) can provide straightforward supervision, it would treat partial labels as one-hot targets, leading to the loss of uncertainty information and the risk of reinforcing noisy signals. In order to fully exploit the informative consensus partial labels, in \ourmethod, we design two complementary loss functions for classifier training: the \textit{vote-aggregated cross-entropy (VaCE) loss} on the label space and the \textit{partial label similarity (PaSim) loss} on the representation space.

\subsubsection{\textbf{Classification Model}}
In this module, we first initialize the parameters randomly to construct a two-layer GCN~\cite{kipf2017semi} as the classification model. Specifically, the node embeddings $\mathbf{H}^{(cls)}$ of all nodes and the prediction probability matrix $\mathbf{P}^{(cls)}$ of the classification model are obtained by:

\begin{equation}
\begin{cases}{}
\mathbf{H}^{(cls)} = \sigma(\mathbf{\hat{A}} \mathbf{X} \mathbf{W}_{emb}), \\
\mathbf{P}^{(cls)} = \operatorname{softmax}(\sigma(\mathbf{\hat{A}} \mathbf{H}^{(cls)} \mathbf{W}_{prd})),
\end{cases}
\label{eq33}
\end{equation}
where $\mathbf{W}_{emb}$ and $\mathbf{W}_{prd}$ are trainable parameters of the student model, $\mathbf{\hat{A}} = \mathbf{\tilde{D}^{-\frac{1}{2}}} \mathbf{\tilde{A}} \mathbf{\tilde{D}^{-\frac{1}{2}}}$ where $\mathbf{\tilde{A}} = \mathbf{A} + \mathbf{I}$, $\mathbf{\tilde{D}}$ is a diagonal matrix with $\tilde{d}_{ii} = \sum_j{\tilde{a}_{ij}}$ and $\mathbf{I}$ is the identity matrix, $\sigma$ denotes the activation function. 

Note that we adopt such a basic GNN as our classifier to clearly demonstrate the effectiveness of our proposed framework without the interference of complex architectures. In practice, the classification model can be alternatively replaced by any advanced GNNs for node classification, such as GraphSAGE~\cite{hamilton2017inductive} or GAT~\cite{velivckovic2018graph}.

\subsubsection{\textbf{Vote-Aggregated Cross-Entropy (VaCE) Loss}}
Different from one-hot labels or binary partial labels, the consensus partial labels in \ourmethod contain richer information about the class distribution, which is modeled by the annotators’ voting during label ensembling. To fully exploit the distributional information in consensus partial labels, it is crucial to design a classification loss that considers the voting behaviors. 
To this end, we design a \textbf{V}ote-\textbf{a}ggregated \textbf{C}ross-\textbf{E}ntropy (VaCE) loss $\mathcal{L}_{vace}$, an improvement over the conventional cross-entropy, to leverage the collective intelligence of the consensus labels for learning more robust decision patterns. The loss function is defined as:
\begin{eqnarray}
\mathcal{L}_{vace} = -\sum_{i \in \mathcal{T}}\sum_{j=1}^{c} \mathbb{I}(\tilde{y}_{ij} \neq 0) \log \mathbf{p}_{ij}^{(cls)}\cdot\tilde{y}_{ij},
\label{eq8}
\end{eqnarray}
where $\mathcal{T}$ denotes the training set. Different from the conventional cross-entropy loss, our VaCE loss explicitly accounts for the voting tendencies of annotators by assigning greater update strength to categories with higher vote counts. In this way, the model benefits from fine-grained supervision signals and avoids the overconfidence issue of one-hot training.

\subsubsection{\textbf{Partial label Similarity (PaSim) Loss}}
In the field of partial label learning, partial labels of the samples are typically constructed via manual annotation, rule-based generation, or random noise injection, where the non-target classes are often meaningless. Consequently, existing algorithms mainly focus on identifying the correct labels from the partial labels. 
However, the consensus labels constructed in \ourmethod not only contain the class information of nodes, but their non-target classes also carry rich semantic information (\eg, reflecting the node’s position near classification boundaries). To leverage this semantic information to improve model generalization, we design a novel loss function, referred to as the \textbf{Pa}rtial label \textbf{Sim}ilarity (PaSim) loss. Specifically, we first compute the consensus partial label similarity matrix $\mathbf{S}_{lab}$ and the embedding similarity matrix $\mathbf{S}_{emb}$:
\begin{equation}
\mathbf{S}_{lab} = \tilde{\mathbf{Y}} \cdot \tilde{\mathbf{Y}}^\top,\ \mathbf{S}_{emb} = \mathbf{H}^{(cls)} \cdot \mathbf{H}^{(cls)\top},
\label{eq6}
\end{equation}
where $^\top$ denotes the matrix transpose operation., and $\cdot$ represents matrix multiplication. By calculating the pair-wise similarity of partial labels, the consensus partial label similarity matrix $\mathbf{S}_{lab}$ captures the semantic affinity information of nodes in the partial label space. Such a semantic affinity can serve as a valuable supervision signal for guiding the embedding distribution in the representation space. 
Subsequently, we constrain the node distribution in the classifier’s representation space to be consistent with the label space through the following consistency constraint:
\begin{equation}
\mathcal{L}_{{pasim}} = \| \mathbf{S}_{emb} - \mathbf{S}_{lab} \|_1 = \frac{1}{n^2} \sum_{i=1}^n \sum_{j=1}^n \left| s^{emb}_{ij} - s^{lab}_{ij} \right|.
\label{eq7}
\end{equation}

Finally, by combining the VaCE loss and PaSim loss, we obtain the overall partial learning objective for the classification model:
\begin{eqnarray}
\mathcal{L} = \mathcal{L}_{vace} + \lambda \mathcal{L}_{pasim},
\label{eq9}
\end{eqnarray}
where $\lambda$ is a hyperparameter that controls the weight of the PaSim loss. The two losses constrain the classifier in different spaces (i.e., the label space and the representation space) and leverage the class distribution and similarity information of consensus partial labels, and hence complement each other to enhance model robustness and discriminability.


\subsection{Iterative Self-Training Framework}\label{sec:enhance}

Although the two modules in \ourmethod can realize the annotation and utilization of partial labels, if they are executed in a purely sequential manner, the overall effectiveness of \ourmethod may be undermined by noisy labels. Concretely, the low-quality labels from the original data may affect the annotation performance, leading to inaccurate partial labels. Then, the sub-optimal partial labels can further hinder the training of the classification model. 

To alleviate the impact of original noisy labels and improve learning robustness, we employ an iterative self-training strategy \cite{li2018deeper} for \ourmethod, where the annotation module and the partial label learning module are jointly optimized in a closed-loop and bootstrapping manner. In each iteration, we use the annotated partial labels to update the partial label learning module, and then use the predictions generated by the partial label learning module to further refine the annotators. Running in the loop with $\tau$ iterations, both modules are progressively refined and mutually enhanced, resulting in better partial labels and classification results. 

In order to form a closed loop, a label enhancement strategy is needed to bridge the predictions of the classifier back to the annotator. In \ourmethod, we use a simple yet effective label enhancement strategy that consists of two steps: \textit{noisy label filtering} and \textit{pseudo-label expansion}. The label enhancement strategy focuses on the refinement of training node sets $\mathcal{S}$ and the corresponding labels for annotator training, where $\mathcal{S}$ is initialized as the original labeled node set $\mathcal{V}_L$.


\subsubsection{\textbf{Noisy Label Filtering}}
First, we filter out noisy labels in $\mathcal{S}$ to reduce incorrect supervisory information. Specifically, after the classification model has converged after training with partial labels, we compute the loss set $\mathbf{L}$ of the labeled nodes for each class (e.g., the $j$-th class) based on its predictions:
\begin{equation}
\mathbf{L}_j = \{ l_i \mid y_i = j \}, \quad \text{where} j \in [1,..., c],
\label{eq10}
\end{equation}
where $l_i$ denotes the conventional cross-entropy loss of node $v_i$ given class $j$. 

Next, following the small-loss strategy, nodes with higher loss are more likely to contain noisy labels \cite{gui2021towards,lu2022selc}. Therefore, we compute a noisy loss threshold using the following equation to select the nodes with the top $r$\% highest loss values within each class:
\begin{eqnarray}
\boldsymbol{\varphi}_{i} = sort(\mathbf{L}_i)_{\lceil {\sigma}_i r \rceil},
\label{eq11}
\end{eqnarray}
where ${\sigma}_i$ denotes the number of labelled nodes of class $i$, $r$ is a hyper-parameter and $\boldsymbol{\varphi}_{i}$ is the noisy loss threshold for class $i$. Finally, we regard nodes with loss values exceeding this threshold as noisy nodes and remove their labels. In subsequent training, these nodes are treated as unlabeled nodes. By executing the noisy label filtering, the unreliable nodes in the previous $\mathcal{S}$ will be removed.

\subsubsection{\textbf{Pseudo-Label Expansion}}
After filtering out noisy labels in the dataset, we increase the supervisory information by assigning pseudo-labels to the unlabeled nodes. Specifically, we first compute the predicted classes $y^{(cls)}$ and confidence $\delta_i$ of all unlabeled samples using the classification model:
\begin{equation}
\begin{cases}{}
y^{(cls)}_i = \arg\max(p^{(cls)}_i), \\
\delta_i = \max(p^{(cls)}_i).
\end{cases}
\label{eq12}
\end{equation}

Subsequently, we assign pseudo labels to the top $\alpha$ ($\alpha$ is a hyper-parameter) nodes with the highest prediction confidence in each class. These pseudo labels are then added to $\mathcal{S}$ to further optimize the annotators. To sum up, the pseudo-label expansion enables us to exploit high-confidence predictions to enrich the supervision signals and improve annotation quality.

After the two-step label enhancement process, the training node set $\mathcal{S}$ for the annotation module can be refined with cleaner and more informative supervision signals. Then, we can continue training the predictor of each annotator in the next iteration, which improves both annotation quality and then improves the classifier performance. 

\subsection{Complexity Analysis}\label{appe:complex}
\textbf{Time Complexity.}
The computational overhead of PaSta primarily stems from its iterative self-training process. Notably, the graph convolutional feature extractors of the annotators are pre-trained once and then frozen, which significantly reduces ongoing training costs. In each self-training iteration, the computational cost of PaSta is dominated by three components: (1) Forward propagation through $t$ frozen GCN annotators to obtain node representations, costing O($t \cdot n^2d$), where $t$, $n$ and $d$ are the number of annotators, nodes and feature dimensions, respectively; (2) Training $t$ lightweight predictors (fully connected layers) on the labeled nodes, requiring O($t \cdot \mathcal{T} \cdot d \cdot c$), where $\mathcal{T}$ and $c$ are the number of training nodes and classes, respectively; and (3) Training the main GCN classifier with dual-space losses, where the GCN forward pass and PaSim loss computation both cost O($n^2d$), and the VaCE loss adds O($\mathcal{T} \cdot c$). Here, the graph convolution and similarity computation are the dominant factors in the overall complexity. As a result, the overall time complexity over $\tau$ iterations is O($\tau \cdot t \cdot n^2d$). Obviously, it is quadratic to the node number $n$, differing from standard GNN training only in the constant factors.

\textbf{Space Complexity.}
The space consumption of \ourmethod is dominated by model parameters and intermediate computation during dual-space optimization. The parameter storage includes $t$ frozen GCN feature extractors ($O(t \cdot d^2)$), $t$ lightweight predictors ($O(t \cdot d \cdot c)$), and the main GCN classifier ($O(d^2)$). Beyond parameters, storing node embeddings incurs $O(nd)$ space, while the similarity matrices required for the PaSim loss dominate memory with $O(n^2)$ complexity to capture pairwise semantic affinities. Overall, \ourmethod has a total space complexity of $O(n^2 + t \cdot d^2)$, scaling quadratically with the number of nodes $n$ due to dense similarity matrices and multiple GCN annotators.


\section{Experiments}
\renewcommand\arraystretch{1.2}
\tabcolsep = 0.08cm
\begin{table*}[!t]
  \centering
  \caption{Classification accuracy (\%) of all methods under different noise rates on four datasets. Best results are in bold.}
  \small

  \begin{minipage}[t]{0.48\textwidth}
    \centering
    \textbf{Cora} \\
    \vspace{2pt}
    \begin{tabular}{c c c c c c}
      \cline{1-6}
      \multirow{2}{*}{\centering Methods} & \multicolumn{3}{c}{Uniform} & \multicolumn{2}{c}{Pair} \\
              & 20\% & 40\% & 60\% & 20\% & 40\% \\
      \cline{1-6}
            GCN     & 74.8 (0.9) & 61.5 (0.3) & 36.1 (1.8) & 69.4 (0.6) & 53.6 (0.9) \\
          GAT     & 76.4 (0.7) & 61.6 (4.4) & 36.2 (2.9) & 72.1 (0.7) & 56.6 (2.5) \\
          \cline{1-6}
          DGI     & 82.3 (0.2) & 70.2 (0.2) & 46.9 (0.2) & 78.2 (0.2) & 65.3 (0.3) \\
          GCA     & 78.9 (0.7) & 76.7 (0.7) & 50.4 (3.7) & 78.3 (1.0) & 68.9 (1.7) \\
          SUGRL   & 83.2 (0.4) & 70.5 (0.1) & 43.1 (0.2) & 78.0 (0.2) & 64.3 (0.6) \\
          \cline{1-6}
          JoCoR   & 76.0 (1.0) & 63.5 (2.2) & 38.9 (2.7) & 71.6 (2.0) & 54.9 (0.9) \\
          NRGNN   & 79.6 (0.4) & 74.5 (1.4) & 45.8 (2.5) & 75.3 (0.7) & 65.4 (2.1) \\
          MTS-GNN & 79.5 (1.2) & 76.9 (1.4) & 60.1 (3.8) & 78.9 (0.9) & 73.2 (2.0) \\
          BO-NNC & 82.6 (1.1) & 81.1 (0.8) & 70.3 (2.2) & 80.8 (1.2) & 78.3 (0.9) \\
          \cline{1-6}
          \textbf{\ourmethod} & \textbf{83.4 (0.6)} & \textbf{82.0 (0.4)} & \textbf{71.9 (1.2)} & \textbf{81.0 (0.4)} & \textbf{78.7 (0.5)} \\
      \cline{1-6}
    \end{tabular}
  \end{minipage}
  \hfill
  \begin{minipage}[t]{0.48\textwidth}
    \centering
    \textbf{CiteSeer} \\
    \vspace{2pt}
    \begin{tabular}{c c c c c c}
      \cline{1-6}
      \multirow{2}{*}{\centering Methods} & \multicolumn{3}{c}{Uniform} & \multicolumn{2}{c}{Pair} \\
              & 20\% & 40\% & 60\% & 20\% & 40\% \\
      \cline{1-6}
            GCN     & 63.1 (1.0) & 59.8 (0.5) & 38.3 (0.6) & 61.7 (0.2) & 51.7 (0.7) \\
          GAT     & 64.4 (1.1) & 60.8 (2.0) & 41.5 (1.4) & 64.5 (1.7) & 51.5 (1.9) \\
          \cline{1-6}
          DGI     & 66.6 (0.2) & 68.2 (0.3) & 45.9 (0.1) & 69.2 (0.1) & 54.5 (0.3) \\
          GCA     & 54.9 (0.6) & 53.3 (1.0) & 33.6 (2.2) & 58.6 (0.7) & 53.3 (0.6) \\
          SUGRL   & 59.9 (0.3) & 55.3 (0.3) & 19.8 (2.2) & 64.8 (0.1) & 40.9 (0.2) \\
          \cline{1-6}
          JoCoR   & 65.2 (2.6) & 63.4 (1.1) & 39.6 (5.1) & 65.0 (2.3) & 51.4 (2.4) \\
          NRGNN   & 67.4 (1.2) & 60.6 (1.8) & 46.5 (1.9) & 66.9 (2.8) & 53.1 (2.3) \\
          MTS-GNN & 72.3 (2.3) & 68.9 (0.7) & 57.1 (1.9) & 69.1 (1.5) & 58.2 (3.0) \\
          BO-NNC & 70.1 (0.3) & 69.6 (0.8) & 67.1 (1.1) & 72.2 (0.8) & 66.4 (1.5) \\
          \cline{1-6}
          \textbf{\ourmethod} & \textbf{72.7 (1.0)} & \textbf{71.2 (0.3)} & \textbf{69.3 (0.8)} & \textbf{72.2 (1.2)} & \textbf{67.6 (1.4)} \\

      \cline{1-6}
    \end{tabular}
  \end{minipage}

  \vspace{0.5cm} 

  \begin{minipage}[t]{0.48\textwidth}
    \centering
    \textbf{DBLP} \\
    \vspace{2pt}
    \begin{tabular}{c c c c c c}
      \cline{1-6}
      \multirow{2}{*}{\centering Methods} & \multicolumn{3}{c}{Uniform} & \multicolumn{2}{c}{Pair} \\
              & 20\% & 40\% & 60\% & 20\% & 40\% \\
      \cline{1-6}
            GCN     & 80.3 (0.4) & 77.5 (1.2) & 59.9 (3.9) & 78.2 (1.0) & 60.2 (2.8) \\
          GAT     & 81.2 (0.2) & 80.1 (0.4) & 61.4 (1.2) & 80.1 (0.5) & 68.7 (1.4) \\
          \cline{1-6}
          DGI     & 80.5 (0.1) & 78.8 (0.2) & 75.7 (0.1) & 80.3 (0.2) & 67.4 (1.1) \\
          GCA     & 83.5 (0.1) & 82.5 (0.4) & 66.7 (3.0) & 82.7 (0.5) & 70.3 (2.4) \\
          SUGRL   & 78.5 (0.1) & 77.4 (0.3) & 60.9 (0.8) & 76.0 (0.3) & 62.4 (1.2) \\
          \cline{1-6}
          JoCoR   & 80.6 (0.3) & 78.2 (0.9) & 60.7 (3.5) & 78.7 (0.8) & 61.6 (2.4) \\
          NRGNN   & 80.6 (0.6) & 81.7 (0.4) & 68.7 (2.0) & 80.3 (0.9) & 69.5 (2.6) \\
          MTS-GNN & 82.3 (0.3) & 82.2 (0.3) & 72.0 (3.0) & 83.4 (0.2) & 79.0 (0.3) \\
          BO-NNC & 83.6 (0.1) & 83.4 (0.8) & 79.6 (0.5) & 83.6 (0.2) & 77.8 (1.3) \\
          \cline{1-6}
          \textbf{\ourmethod} & \textbf{83.9 (0.1)} & \textbf{83.6 (0.1)} & \textbf{80.8 (0.5)} & \textbf{83.7 (0.1)} & \textbf{79.2 (0.6)} \\

      \cline{1-6}
    \end{tabular}
  \end{minipage}
  \hfill
  \begin{minipage}[t]{0.48\textwidth}
    \centering
    \textbf{Computers} \\
    \begin{tabular}{c c c c c c}
      \cline{1-6}
      \multirow{2}{*}{\centering Methods} & \multicolumn{3}{c}{Uniform} & \multicolumn{2}{c}{Pair} \\
              & 20\% & 40\% & 60\% & 20\% & 40\% \\
      \cline{1-6}
            GCN     & 84.9 (0.6) & 83.5 (0.6) & 77.9 (0.3) & 84.0 (0.4) & 71.2 (1.7) \\
          GAT     & 84.5 (0.6) & 80.6 (2.0) & 76.0 (1.1) & 84.7 (1.0) & 75.6 (1.1) \\
          \cline{1-6}
          DGI     & 81.5 (0.1) & 79.4 (0.2) & 75.9 (0.3) & 79.6 (0.1) & 69.2 (0.9) \\
          GCA     & 78.0 (0.6) & 74.7 (0.9) & 63.6 (0.5) & 80.9 (0.2) & 69.7 (1.8) \\
          SUGRL   & 85.0 (0.1) & 83.1 (0.1) & 78.5 (0.2) & 84.5 (0.1) & 74.0 (0.4) \\
          \cline{1-6}
          JoCoR   & 85.2 (0.8) & 84.0 (0.4) & 77.9 (0.6) & 84.4 (0.9) & 75.4 (1.3) \\
          NRGNN   & 85.2 (0.9) & 84.1 (0.9) & 77.5 (0.8) & 84.3 (0.3) & 75.0 (1.6) \\
          MTS-GNN & 86.1 (0.6) & 83.6 (2.2) & 80.2 (1.2) & \textbf{86.7 (0.3)} & 79.7 (1.7) \\
          BO-NNC  & 85.6 (0.6) & 84.4 (0.6) & 80.3 (0.7) & 84.1 (0.8) & 80.0 (3.4) \\
          \cline{1-6}
          \textbf{\ourmethod} & \textbf{86.5 (0.3)} & \textbf{84.8 (0.2)} & \textbf{80.7 (0.6)} & 85.6 (0.3) & \textbf{83.5 (0.7)} \\

      \cline{1-6}
    \end{tabular}
  \end{minipage}

  \label{table:classification_accuracy}
\end{table*}

\subsection{Experimental Settings}

\subsubsection{\textbf{Datasets}}
We evaluate the proposed method on five real-world graph datasets: three citation networks (Cora and Citeseer~\cite{yang2016revisiting}), an academic collaboration network (DBLP~\cite{bojchevski2018deep}), and two product co-purchase network (Computers and Photo~\cite{shchur2018pitfalls}).
The detailed statistics of the datasets used in this work are provided in Appendix B.

\subsubsection{\textbf{Comparison Methods}}
The following comparison methods are considered for a comprehensive comparison: two classical GNN architectures (GCN~\cite{kipf2017semi} and GAT~\cite{velivckovic2018graph}), three representative self-supervised graph learning approaches (DGI~\cite{velivckovic2018deep}, GCA~\cite{zhu2021graph}, and SUGRL~\cite{mo2022simple}), and four state-of-the-art noisy node classification methods (JoCoR~\cite{wei2020combating}, NRGNN~\cite{dai2021nrgnn}, MTS-GNN~\cite{liu2023multi}, and BO-NNC~\cite{liu2025noisy}). This selection encompasses traditional supervised models, label-efficient self-supervised paradigms, and specialized noise-robust techniques, providing a thorough evaluation across diverse methodological categories. 

\subsubsection{\textbf{Experimental Setup}}
In our experiments, we follow previous studies~\cite{han2018co,jiang2018mentornet} to partition each dataset into three disjoint subsets: training, validation, and test. For Cora and Citeseer, we adopt the standard splits provided by the PyTorch Geometric library. For DBLP, Computers and Photo, we follow~\cite{liu2023multi}, randomly selecting 5\%, 10\%, and 60\% of the nodes for the training, validation, and test sets, respectively.
In addition, following~\cite{han2018co,jiang2018mentornet}, we inject noise into the datasets with different noise ratios. We consider two types of noise, namely \textit{uniform noise} and \textit{pair noise}: 
\begin{itemize}[leftmargin=*]
\item[$\bullet$]\textbf{Uniform noise:} Each node label is randomly flipped to any other class with equal probability.

\item[$\bullet$]\textbf{Pair noise:} Node labels are flipped to predefined paired classes according to specified class mappings.
\end{itemize}

\subsection{Result Analysis}
We assess the effectiveness of \ourmethod by comparing it against all comparison methods across five datasets under different noise levels. The results on four representative datasets are summarized in Table \ref{table:classification_accuracy}, and the results on the remaining dataset, Photo, are provided in Appendix A.

From the results, we have the following observations. 
\ding{182}~The proposed \ourmethod consistently achieves the best performance across four datasets. For instance, \ourmethod achieves an average improvement of 1.1\% over the strongest competitor (BO-NNC) and 12.7\% over the weakest one (GCN). 
The reason is that the proposed VaCE loss and PaSim loss enable the classifier to effectively learn the correct classification patterns from consensus labels, thereby preventing overfitting to noisy labels in the dataset and achieving better generalization performance. 
\ding{183}~\ourmethod shows substantial advantages over all semi-supervised and self-supervised node classification methods (\ie GCN, GAT, DGI, GCA, and SUGRL). For example, compared with DGI, which achieves the highest performance among the four methods, \ourmethod achieves average improvements of 8.3\% under uniform noise and 8.5\% under pair noise across all datasets. 
This is because \ourmethod effectively enhances the quality of label information in the dataset through noise filtering and label enrichment, providing more reliable supervision for model training. 
\ding{184}~\ourmethod also outperforms existing NNC methods (\eg JoCoR, NRGNN, MTS-GNN, and BO-NNC). Compared to BO-NNC, \ourmethod achieves the highest performance among these four methods, by an average of 1.1\% across all experimental conditions. 
The reason is that \ourmethod mitigates the limitations of existing NNC methods by iteratively optimizing the annotation module and the partial label learning module, thereby achieving stronger robustness against label noise. This also demonstrates that introducing partial labels to tackle the NNC problem is reasonable and effective.

\subsection{Ablation Study}
\ourmethod has four key designs, namely annotators to construct consensus labels, the VaCE loss, the PaSim loss, and a self-training strategy. To evaluate the contribution of each component, we conducted extensive experiments under different settings, including uniform noise with a rate of 60\% and pair noise with a rate of 40\%. 

First, we evaluated the performance of different combinations of the PaSim loss (C1), VaCE loss (C2), and self-training strategy (C3). The results are summarized in Table \ref{tab:ablation1}. We can observe that: \ding{182}~Our method with all components improves by 4.2\% on average compared to the versions with any two components. This suggests that all three components are essential to the overall performance of our method, confirming the soundness of our approach. 
\ding{183}~Compared with the case without using the PaSim loss and VaCE loss, incorporating these two losses yields an average improvement of 1.7\%. This indicates that the proposed losses effectively help the classifier learn correct classification patterns from the consensus labels.
\ding{184}~Performing the self-training strategy results in an average improvement of 9.6\% compared to methods that do not utilize it. This demonstrates that the designed iterative optimization process can effectively combine the strengths of one-hot label learning and partial label learning, thereby enabling the student model to achieve stronger generalization performance.

\begin{figure}[!t]
    \centering
    \subfigure[Uniform noise (60\%)]{
    \begin{minipage}[t]{0.44\linewidth}
    \centering
    \includegraphics[width=\linewidth, trim = 0 0 865 0, clip]{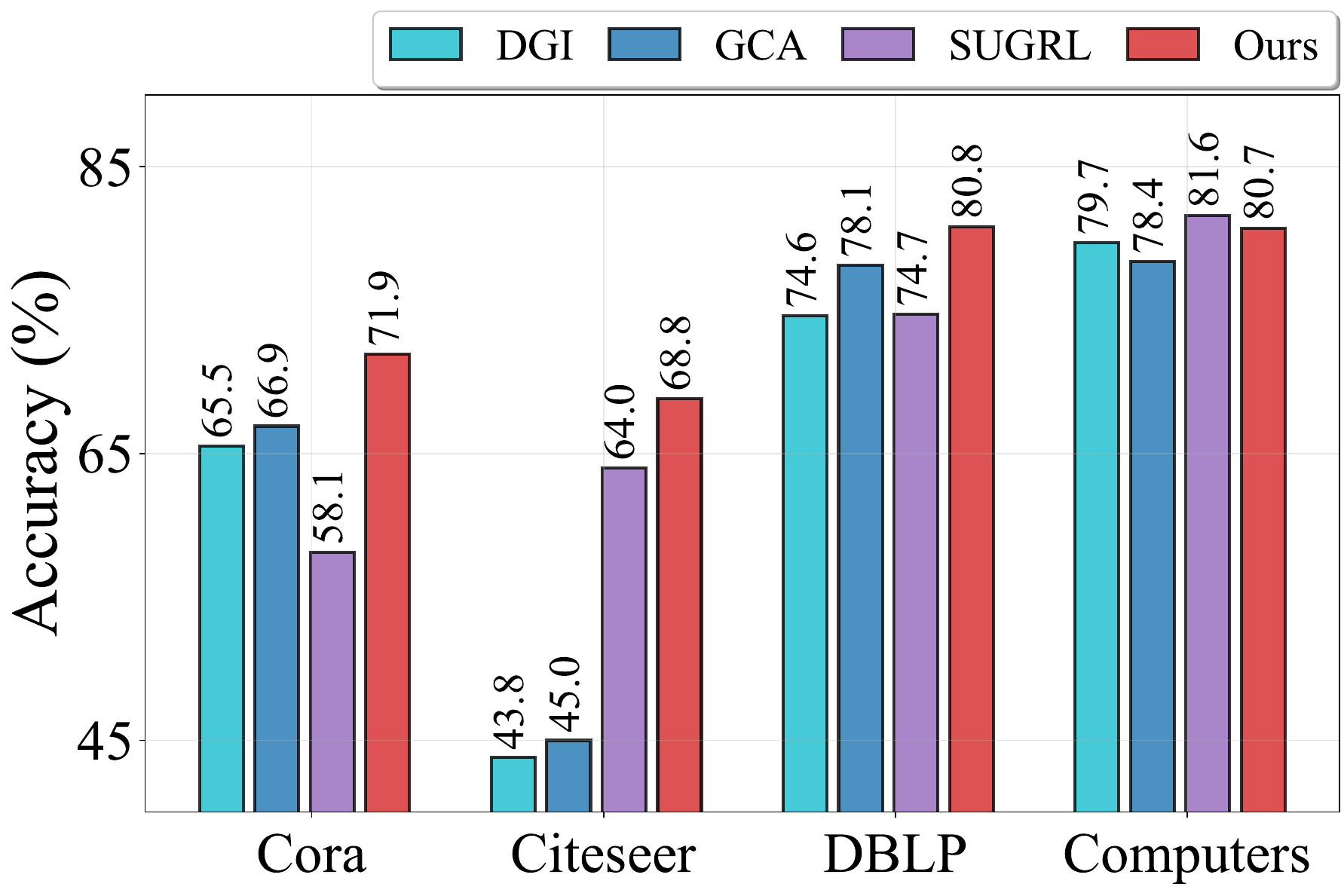}
    \end{minipage}
    }
    \hspace{6pt}
    \subfigure[Pair noise (40\%)]{
    \begin{minipage}[t]{0.44\linewidth}
    \centering
    \includegraphics[width=\linewidth, trim = 865 0 0 0, clip]{figure/Ablation.pdf}
    \end{minipage}
    }
    \caption{Node classification performance of \ourmethod with different annotators under the highest noise rates for uniform and pair noise.}
    \label{fig:ablation2}
\end{figure}

Second, we evaluated the contribution of annotators to construct consensus labels. Specifically, we compared the performance of our method with that of using a single annotator, where partial labels are constructed by selecting the top-2 predicted classes with the highest confidence. The results are shown in Figure \ref{fig:ablation2}. The experimental results demonstrate that the proposed strategy of collaboratively constructing partial labels with multiple annotators significantly outperforms the approach that relies on a single annotator. Specifically, compared with using DGI, GCA, and SUGRL individually to generate partial labels, our method achieves an average improvement of 7.5\%, 6.9\%, and 6.1\% in classification accuracy, respectively. These results verify the effectiveness of the multi-annotator strategy, which fully exploits the complementarity of different annotators’ predictions to substantially enhance the quality of partial labels, thereby further improving the generalization performance of the classifier.

\renewcommand\arraystretch{1.1} 
\tabcolsep = 0.14cm 
\begin{table}[!t]
  \centering
  \caption{Classification accuracy (\%) of \ourmethod with different component combinations under the highest noise rates for uniform and pair noise.}
  \small
  {
    \begin{tabular}{ccc cccc}
    \multicolumn{7}{c}{Uniform noise (60\%)}\\
    \cline{1-7}
    C1 & C2 & C3 & Cora & Citeseer & DBLP & Computers\\
    \cline{1-7}
    \checkmark &            &            & 59.9 (3.1) & 36.0 (11.3) & 71.5 (1.5) & 79.3 (0.4) \\
               & \checkmark &            & 62.6 (1.7) & 39.9 (9.3)  & 74.1 (1.4) & 79.4 (0.4) \\
               &            & \checkmark & 70.8 (1.0) & 64.0 (6.2)  & 80.3 (0.4) & 80.7 (0.3) \\
    \checkmark & \checkmark &            & 63.1 (1.9) & 38.4 (12.7) & 73.6 (1.5) & 79.6 (0.4) \\
    \checkmark &            & \checkmark & 70.7 (1.7) & 64.6 (4.2)  & 80.3 (0.4) & 80.8 (0.3) \\
               & \checkmark & \checkmark & 70.3 (2.4) & 64.5 (8.5)  & 80.3 (0.5) & 80.7 (0.7) \\
               \cline{1-7}
    \checkmark & \checkmark & \checkmark & \textbf{71.9 (1.2)} & \textbf{69.3 (0.8)}  & \textbf{80.8 (0.5)} & \textbf{80.7 (0.6)} \\
    \cline{1-7}
    \end{tabular}%
    \\
    \rule{0pt}{3pt} \\
    
    \begin{tabular}{ccc cccc}
    \multicolumn{7}{c}{Pair noise (40\%)}\\
    \cline{1-7}
    C1 & C2 & C3 & Cora & Citeseer & DBLP & Computers\\
    \cline{1-7}
    \checkmark &            &            & 73.0 (1.1) & 54.3 (5.1) & 72.1 (0.6) & 77.9 (2.8) \\
               & \checkmark &            & 72.8 (1.2) & 55.0 (3.7) & 73.8 (0.4) & 79.0 (3.9) \\
               &            & \checkmark & 78.0 (1.1) & 64.4 (1.1) & 77.7 (1.2) & 82.6 (1.1) \\
    \checkmark & \checkmark &            & 73.1 (1.1) & 54.9 (3.8) & 73.1 (0.3) & 79.5 (3.4) \\
    \checkmark &            & \checkmark & 77.7 (0.8) & 64.2 (1.3) & 76.6 (1.4) & 82.7 (0.6) \\
               & \checkmark & \checkmark & 78.5 (0.8) & 67.0 (1.1) & 77.9 (1.2) & 82.5 (0.7) \\
               \cline{1-7}
    \checkmark & \checkmark & \checkmark & \textbf{78.7 (0.5)} & \textbf{67.6 (1.4)}  & \textbf{79.2 (0.6)} & \textbf{83.5 (0.7)} \\
    \cline{1-7}
    \end{tabular}%
  }
  \label{tab:ablation1}
\end{table}

\subsection{Parameter Sensitivity Analysis}
In this subsection, we investigate the sensitivity of \ourmethod to the selection of iterative optimization rounds $\tau$. We vary $\tau$ from \{0, 1, \dots, 5\}, and report the results in Figure~\ref{fig:PS1}. Sensitivity to the loss balance hyperparameter $\lambda$ is provided in Appendix A.

First, according to the Figure~\ref{fig:PS1}, we can see that: \ding{182}~The initial few iterations bring significant performance improvement compared to the non-iterative setting. This is because the self-training mechanism effectively refines the annotators with more reliable supervision, leading to better partial labels and more robust classifier training. \ding{183}~The model performance improves as $\tau$ increases and tends to stabilize when $\tau$ exceeds 3. The reason is that the self-training strategy in our method enables the annotators and the classifier to mutually enhance each other, thereby improving classification performance, and the training converges after approximately three iterations.

\begin{figure}[!t]
    \centering
    \subfigure[Uniform noise (60\%)]{
    \begin{minipage}[t]{0.44\linewidth}
    \centering
    \includegraphics[width=\linewidth, trim = 0 10 710 0, clip]{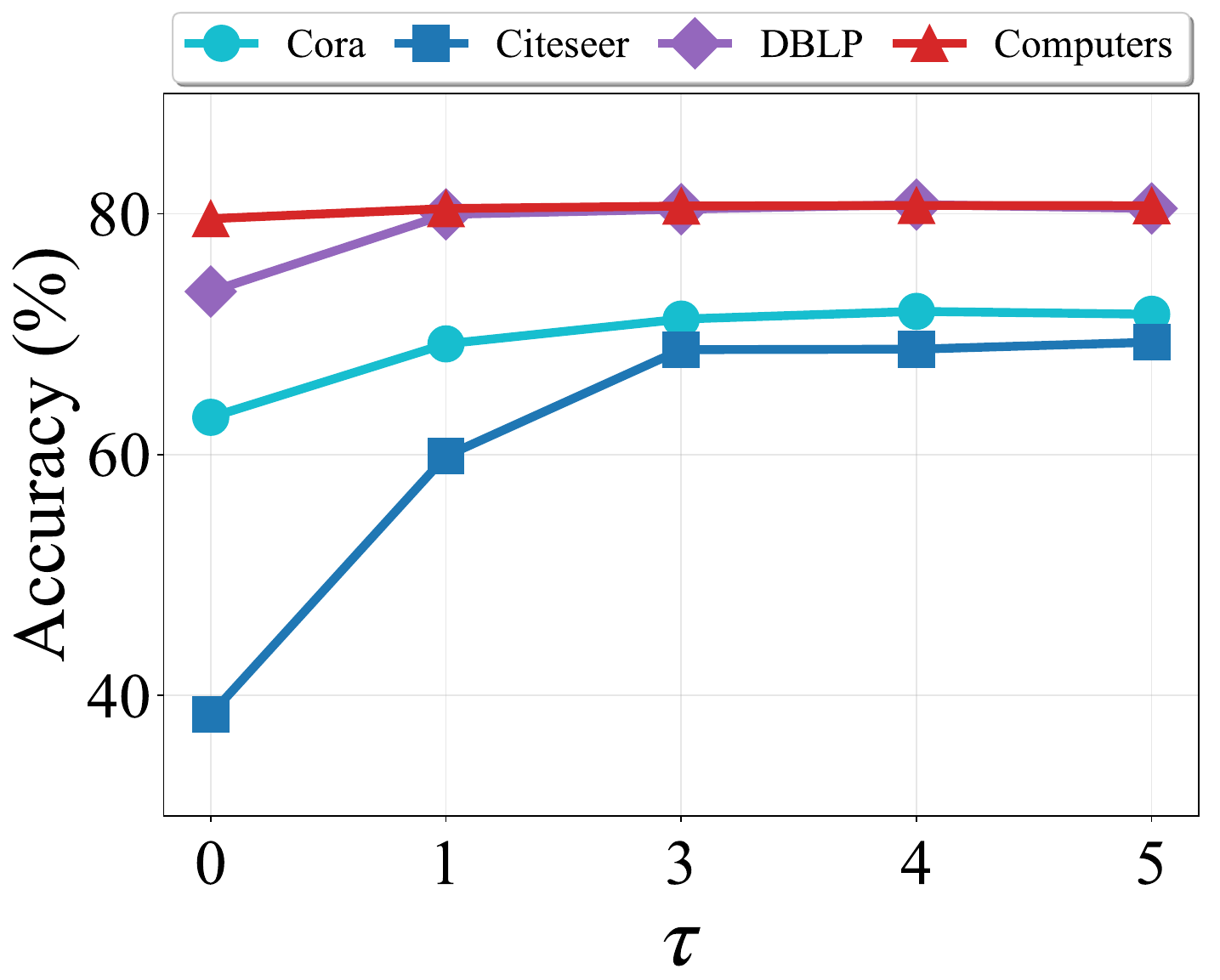}
    \end{minipage}
    }
    \hspace{6pt}
    \subfigure[Pair noise (40\%)]{
    \begin{minipage}[t]{0.44\linewidth}
    \centering
    \includegraphics[width=\linewidth, trim = 710 10 0 0, clip]{figure/PS_T.pdf}
    \end{minipage}
    }
    \caption{Sensitivity Analysis of the Parameter $\tau$ in \ourmethod.}
    \label{fig:PS1}
\end{figure}

\begin{figure}[!t]
    \centering
    \includegraphics[width=0.3\textwidth]{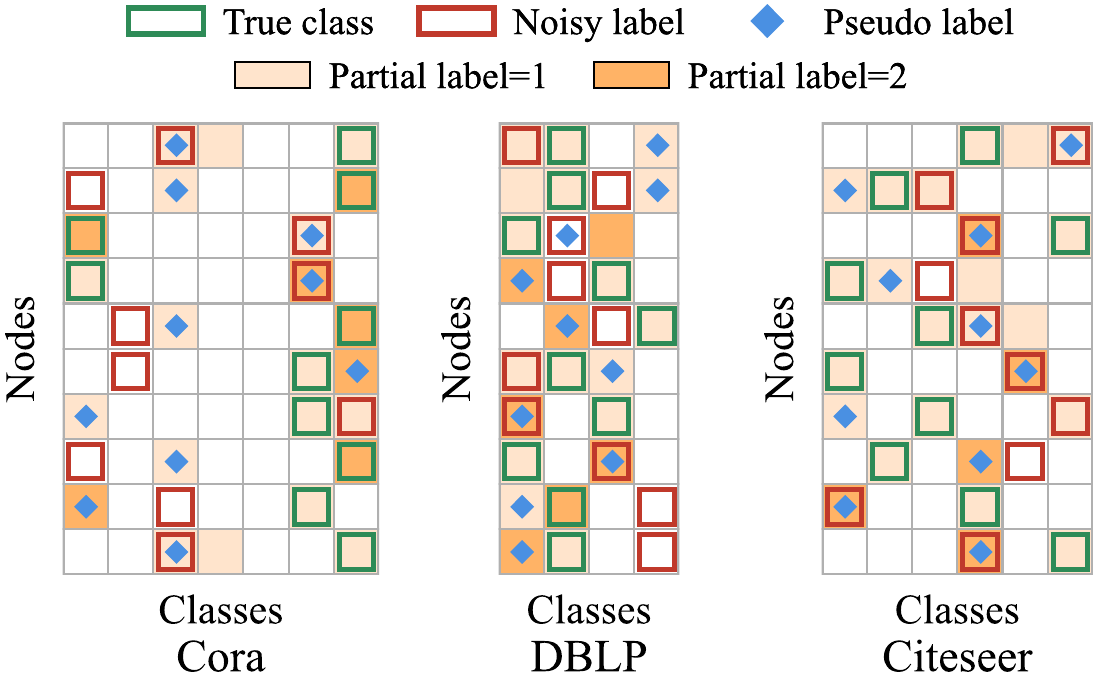}
    \caption{Visualization of consensus partial label and one-hot pseudo-label (ten sampled noisy nodes).}
    \label{fig:visualization}
\end{figure}

\subsection{Consensus Partial Label Visualization}
To intuitively illustrate the difference between the consensus partial labels constructed in this work and the traditional one-hot pseudo-labels, we select ten nodes with noisy labels from each of the three datasets for label visualization. In this experiment, one-hot pseudo-labels were generated by summing the predictions of the three annotators and selecting the class with the highest confidence.
As shown in Figure \ref{fig:visualization}, although some annotators can correctly predict these nodes, the traditional one-hot pseudo-labels based on aggregated probabilities may still assign incorrect classes. In contrast, the consensus partial labels given by \ourmethod can effectively prevent the loss of accurate information by retaining multiple candidate categories, thereby providing more reliable supervisory signals. 

\section{Conclusion}
As noisy labels are commonly encountered in real-world graph data, noisy node classification has become a crucial problem for advancing the practical use of graph-structured data. To address this challenge, we propose a novel \textbf{Pa}rtial label-based \textbf{S}elf-\textbf{t}raining fr\textbf{a}mework (\ourmethod), which breaks the limitations of conventional one-hot label supervision. \ourmethod consists of two synergistic modules: a self-supervised ensemble annotation module for generating high-quality partial labels, and a dual-space partial label learning module that exploits partial labels via complementary loss in both label and representation spaces, and also refines node labels. Through iterative optimization, the modules mutually reinforce each other, progressively improving label quality and classifier robustness. Experiments on five real-world benchmarks show that \ourmethod consistently outperforms existing SOTA methods across different datasets and noise levels.

\bibliographystyle{IEEEtran}
\bibliography{references}

\section*{Appendix}
\subsection*{A. Additional Results}\label{appe:more_results}

\subsubsection*{Classification Results of Photo Dataset}\label{appe:Photo}

We report the node classification performance of \ourmethod and all comparison methods on the Photo dataset under different noise rates in Table~\ref{exm:Photo}. Obviously, \ourmethod also achieves the best performance on the Photo dataset. Specifically, it surpasses the existing SOTA method BO-NNC by 2.27\% and 2.57\% under the uniform and pair noise settings, respectively. This result further demonstrates the robustness and effectiveness of \ourmethod in handling label noise with varying types and intensities.

\renewcommand\arraystretch{1.2}
\tabcolsep = 0.08cm
\begin{table}[htbp]
  \centering
  \caption{Classification accuracy (\%) of all methods under different noise rates on dataset Photo. Best results are in bold.}
  \small
    \resizebox{0.48\textwidth}{!}{
    \begin{tabular}{c c c c c c}
      \cline{1-6}
      \multirow{2}{*}{\centering Methods} & \multicolumn{3}{c}{Uniform} & \multicolumn{2}{c}{Pair} \\
              & 20\% & 40\% & 60\% & 20\% & 40\% \\
      \cline{1-6}
            GCN     & 89.90 (0.47) & 87.03 (1.21) & 81.07 (2.13) & 85.74 (0.95) & 64.27 (1.66) \\
          GAT     & 89.86 (1.24) & 88.17 (1.03) & 81.24 (1.57) & 87.36 (1.25) & 68.67 (1.35) \\
          \cline{1-6}
          DGI     & 90.22 (0.29) & 85.20 (0.14) & 83.49 (0.83) & 85.97 (0.10) & 63.43 (0.90) \\
          GCA     & 86.91 (0.19) & 82.94 (1.04) & 73.73 (1.76) & 86.03 (1.46) & 77.34 (0.33) \\
          SUGRL   & 91.47 (0.10) & 89.22 (0.20) & 83.25 (0.86) & 89.92 (0.13) & 70.88 (0.71) \\
          \cline{1-6}
          JoCoR   & 89.90 (0.45) & 87.49 (0.64) & 82.08 (1.79) & 86.93 (0.24) & 68.73 (1.65) \\
          NRGNN   & 89.79 (0.79) & 87.45 (1.15) & 81.45 (1.50) & 86.37 (1.44) & 67.11 (2.47) \\
          MTS-GNN & 91.97 (0.23) & 88.48 (0.37) & 87.89 (1.64) & 89.21 (0.42) & 79.78 (5.61) \\
          BO-NNC  & 90.82 (1.10) & 88.78 (2.35) & 87.43 (2.64) & 89.63 (0.16) & 81.58 (2.74) \\
          \cline{1-6}
          \textbf{\ourmethod} & \textbf{92.07 (0.25)} & \textbf{91.01 (0.70)} & \textbf{90.75 (0.90)} & \textbf{91.43 (0.99)} & \textbf{84.92 (2.02)} \\
      \cline{1-6}
    \end{tabular}}
    \label{exm:Photo}
    \end{table}

\subsubsection*{Sensitivity to $\lambda$}\label{appe:lambda}

\begin{figure}[htbp]
    \centering
    \subfigure[Uniform noise (60\%)]{
    \begin{minipage}[t]{0.46\linewidth}
    \centering
    \includegraphics[width=\linewidth, trim = 0 10 710 0, clip]{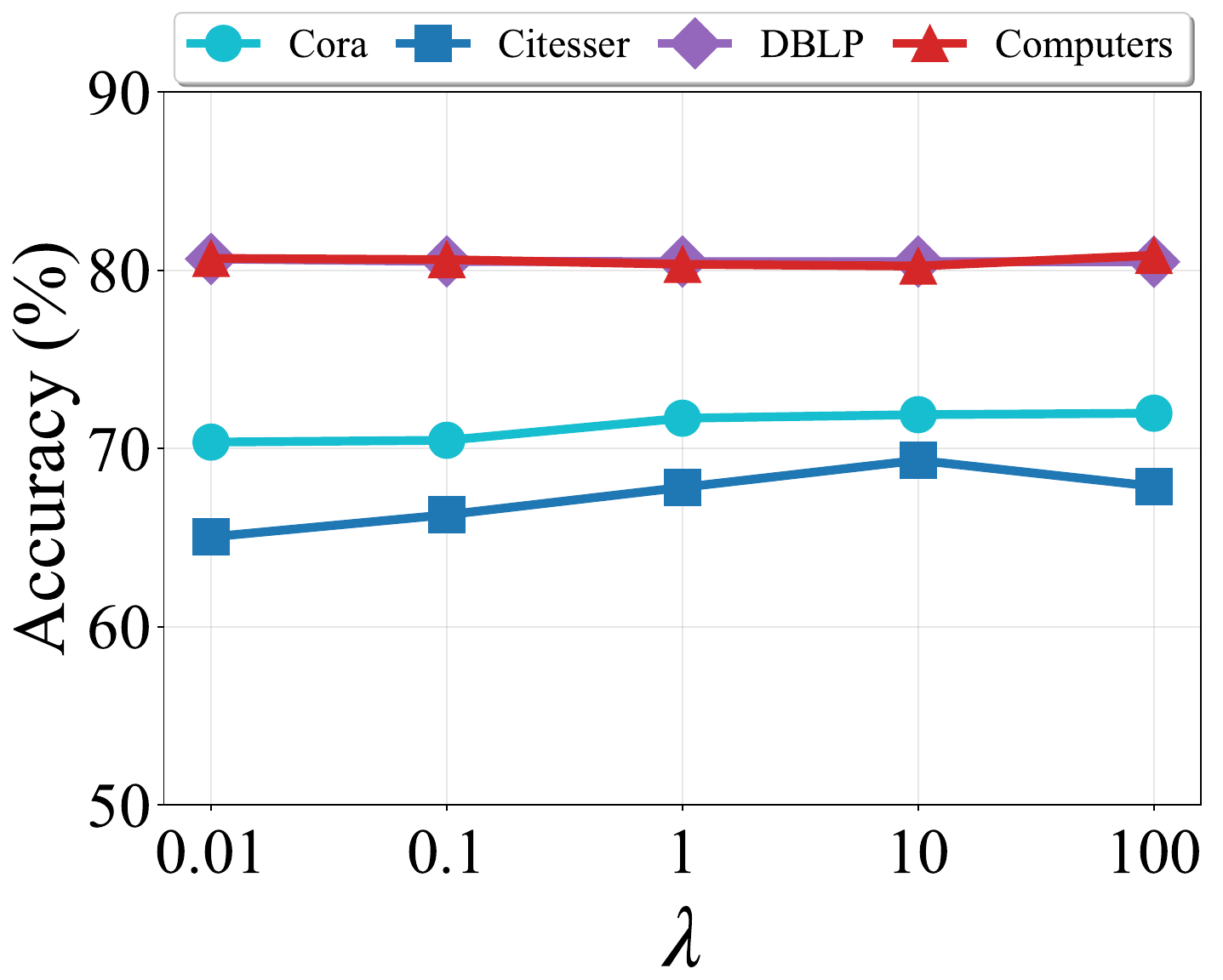}
    \end{minipage}
    }
    \subfigure[Pair noise (40\%)]{
    \begin{minipage}[t]{0.46\linewidth}
    \centering
    \includegraphics[width=\linewidth, trim = 710 10 0 0, clip]{figure/PS_lambda.pdf}
    \end{minipage}
    }
    \caption{Sensitivity Analysis of the Parameter $\lambda$ in \ourmethod.}
    \label{fig:PS2}
\end{figure}

Our method has another key hyper-parameter, i.e., $\lambda$ controlling the weight of the $\mathcal{L}_{pasim}$. To investigate the sensitivity of our method to the hyper-parameter, we vary $\lambda$ in \{0.01, 0.1, 1, 10, 100\}, and report the results in Figure~\ref{fig:PS2}. 
From the figure, we can see that our method is not sensitive to the value of $\lambda$. This is because the optimization objectives of $\mathcal{L}_{pasim}$ and $\mathcal{L}_{vace}$ are inherently aligned, with $\mathcal{L}_{pasim}$ designed to help $\mathcal{L}_{vace}$ achieve its goal more robustly. As a result, the model can effectively optimize both objectives simultaneously regardless of the value of $\lambda$.

\end{document}